\documentclass[conference]{IEEEtran}
\IEEEoverridecommandlockouts

\usepackage{cite}
\usepackage{amsmath,amssymb,amsfonts}
\usepackage{algorithmic}
\usepackage{graphicx}
\usepackage{textcomp}
\usepackage{xcolor}
\usepackage{subcaption}
\usepackage{graphicx}
\usepackage{booktabs}
\usepackage{multirow}
\usepackage{amsmath}
\usepackage{array} 
\def\BibTeX{{\rm B\kern-.05em{\sc i\kern-.025em b}\kern-.08em
    T\kern-.1667em\lower.7ex\hbox{E}\kern-.125emX}}
\begin{document}

\title{MHGNet: Multi-Heterogeneous Graph Neural Network for Traffic Prediction\\
\thanks{*corresponding author: 111124120010@zjut.edu.cn}
}

\author{\IEEEauthorblockN{Mei Wu}
\IEEEauthorblockA{\textit{Hangzhou Dianzi University} \\
Hangzhou, China \\
222320007@hdu.edu.cn}
\and
\IEEEauthorblockN{Yiqian Lin}
\IEEEauthorblockA{\textit{Hangzhou Dianzi University} \\
Hangzhou, China \\
linyq@hdu.edu.cn}
\and
\IEEEauthorblockN{Tianfan Jiang}
\IEEEauthorblockA{\textit{Hangzhou Dianzi University} \\
Hangzhou, China \\
232320005@hdu.edu.cn}
\and
\IEEEauthorblockN{Wenchao Weng*}
\IEEEauthorblockA{\textit{Zhejiang University of Technology} \\
Hangzhou, China \\
 111124120010@zjut.edu.cn}
}

\maketitle

\begin{abstract}
In recent years, traffic flow prediction has played a crucial role in the management of intelligent transportation systems. However, traditional forecasting methods often model non-Euclidean low-dimensional traffic data as a simple graph with single type nodes and edges, failing to capture similar trends among nodes of the same type. This paper proposes MHGNet, a new framework for modeling spatiotemporal multi-heterogeneous graphs, in which the STD Module decouples single-pattern traffic data into multi-pattern traffic data through feature mappings of timestamp embedding matrices and node embedding matrices. Then the Node Clusterer uses the Euclidean distance between nodes and different types of limit points for O(N) time complexity clustering, and the nodes within clusters undergo residual subgraph convolution in the spatiotemporal fusion subgraphs generated by the DSTGG Module before entering the SIE Module for node repositioning and redistribution of weights. This paper conducts a series of ablation and quantitative evaluations on four widely used benchmarks to validate the effectiveness of MHGNet. 
\end{abstract}


\section{Introduction}
The task of traffic prediction (e.g., traffic flow prediction) is based on historical traffic conditions collected from sensors to forecast future traffic conditions\cite{conditions,conditions2,conditions3}. Graph Neural Networks (GNN) have been introduced to effectively manage the non-uniform topological structure of urban traffic networks\cite{research1,research2,research3}. Spatiotemporal Graph Neural Networks (STGNN) combine GNN with various temporal learning methods to capture hidden patterns of spatially irregular signals that change over time, thus addressing the spatiotemporal heterogeneity of non-Euclidean urban data\cite{prediction,prediction2,GCN}. However, existing models still cannot address the following issues in building spatiotemporal relationships between graph nodes:

\begin{figure}[t]
    \centering
    \includegraphics[width=.48\textwidth]{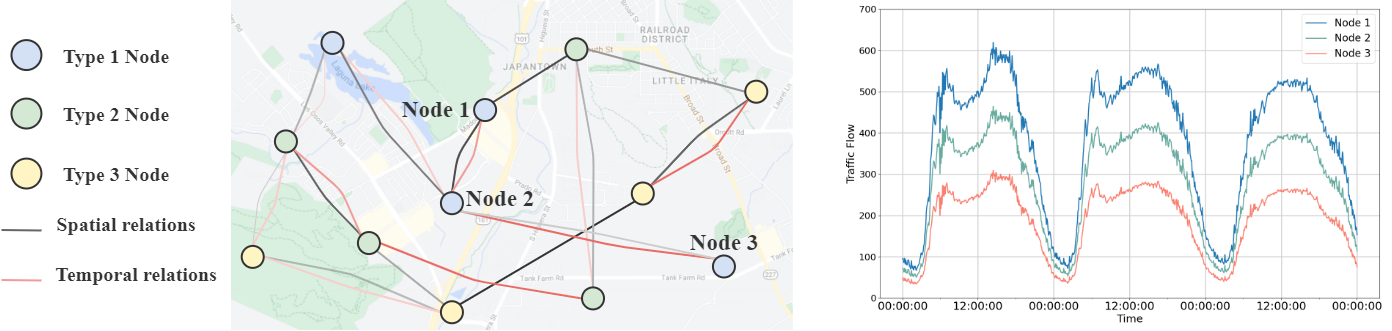}
    \caption{The nodes have different sizes of spatiotemporal relationship similarities. Nodes of the same type have similar traffic patterns.}
    \label{fig1}
\end{figure}

(1) Lack of handling the structural patterns of traffic. Traffic signal features are contributed by multiple traffic patterns (e.g., trucks, passenger cars, broken-down vehicles). As shown in Figure \ref{fig1}, nodes with similar traffic patterns have similar flow (speed) changes, regardless of physical spatial distance.

(2) Lack of handling multi-dimensional spatiotemporal relationships. Current methods typically use Graph Neural Networks (GNN) to aggregate spatial features of single-dimensional flow or speed, and then use RNN or TCN for temporal feature extraction\cite{GCN2,tasks,tasks2}. In reality, true traffic data is dynamically complex, exhibiting multi-dimensional spatiotemporal heterogeneity\cite{AspEm,HEER,COMPGCN} within the spatial area of each node.

(3) Lack of diversity in modeling. Existing methods typically model non-Euclidean traffic data as simple graphs with single-type nodes and edges\cite{networks,networks2,networks3}. However, real-world complex traffic data, as shown in Figure \ref{fig1}, is a multigraph with various types of nodes and edges. The traffic flow of nodes of the same type shows similar trends, and node pairs are connected by two types of edges: temporal and spatial relationships.

To address issues one and two, the Spatiotemporal Decoupling Module in MHGNet is designed to decouple single-pattern traffic data into multi-pattern traffic data through the feature mapping of the timestamp embedding matrix and node embedding matrix, capturing dynamic multi-dimensional spatiotemporal information in the traffic pattern structure. For issue three, the complex traffic network is modeled as a multi-layer heterogeneous graph. A node clustering algorithm is employed, which clusters nodes by calculating the Euclidean distance between the nodes in a multi-dimensional feature space and different types of centroid points. The node sequence numbers are then stored in a node sequence pool.

As shown in Figure \ref{fig2}, the dynamic spatiotemporal graph generation module creates dynamic spatiotemporal subgraphs with weighted spatiotemporal relationships for each node cluster. This enables convolution operations on smaller, simpler homomorphic graphs rather than on large, complex heterogeneous graphs, effectively reducing time complexity and conserving computational resources. Finally, the clustered node IDs from the node sequence pool are used for regressing the node features after subgraph convolution. Two-dimensional convolution and a learnable parameter matrix adjust the weights of the subgraph convolution results at different scales, aggregating the output to iteratively update the sequence dependencies. In summary, the main contributions of this paper are as follows:
\begin{figure}[t]
    \centering
    \includegraphics[width=0.48\textwidth]{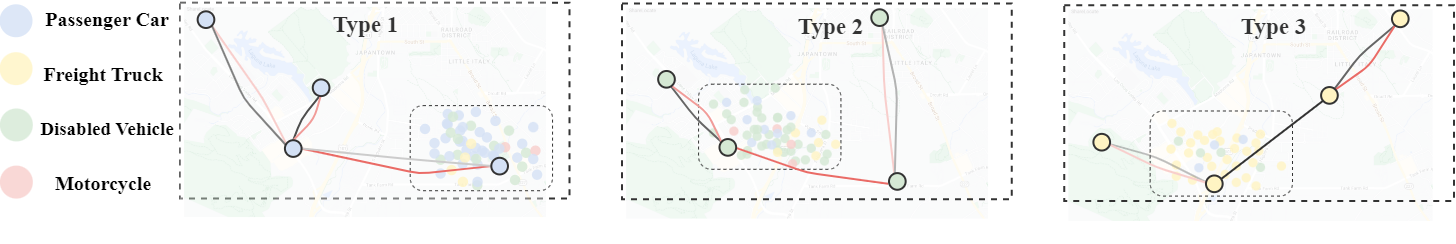}
    \caption{Transforming multi-heterogeneous graph into simple subgraph convolution, with different types of nodes contributing traffic features from different traffic pattern structures.}
    \label{fig2}
\end{figure}
\begin{itemize}
    \item This paper proposes a spatiotemporal modeling method for traffic prediction using heterogeneous multigraphs, mapping single-dimensional traffic data into a multi-dimensional feature space. A node clustering mechanism and a dynamic spatiotemporal graph generation module are employed to convert heterogeneous multigraphs into simple homomorphic graphs. 
    
    \item A subgraph information extraction module is designed to perform gated spatiotemporal information extraction and weight adjustment of convolutional blocks at different scales, with the extracted information regressed back to the original positions of the clustered nodes. 
    
    \item Comprehensive experiments were conducted on two well-known traffic flow datasets and two traffic speed datasets, demonstrating that MHGNet consistently outperforms various competitive baselines.

\end{itemize}

\section{Methodology}
\begin{figure}[t]
    \centering
    \includegraphics[width=0.9\linewidth]{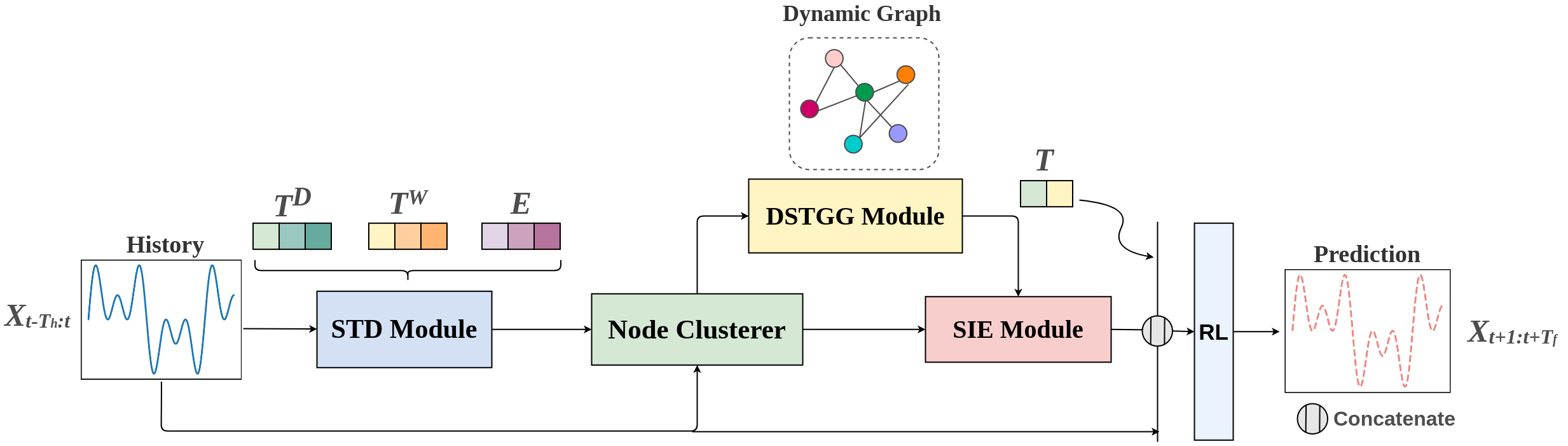}
    \caption{Overall framework diagram (MHGNet)}
    \label{fig3}
\end{figure}

In this section, we will elaborate on the proposed framework and its components. An overview of our model architecture is illustrated in Figure \ref{fig3}.
\subsection{STD Module}
To obtain heterogeneous spatiotemporal feature mappings for different traffic patterns, we introduce daily and weekly periodic feature matrices \( T^D \in \mathbb{R}^{T_h \times N \times D_t} \) and \( T^W \in \mathbb{R}^{T_h \times N \times D_t} \) into the STD Module, and then utilize a node embedding matrix \( E \in \mathbb{R}^{N \times D_s} \) to weight and partition the original traffic flow \( X_{t-T_h:t} \) into \( P \) types of traffic pattern features.

\begin{equation}
\begin{gathered}
    \Omega_{(t,i)}=Sigmoid((ReLU(T_{(t,i)}^D ||T_{(t,i)}^W || E_i ) W_1 ) W_2) \\
    X_n = \hat{X}_{t-T_h:t}\odot \Omega_n \\
    X_p= \hat{X}_{t-T_h:t}-\sum_{n=1}^{p-1}X_n \\
    \end{gathered}
\end{equation}
 Here $\Omega_{(t,i)}$ represents the output proportion of the specific traffic pattern at node $i$ relative to total traffic at time step $t$.
\subsection{Node Clusterer}
\begin{figure}[t]
    \centering
    \begin{subfigure}[b]{0.12\textwidth}
        \centering
        \includegraphics[width=\textwidth]{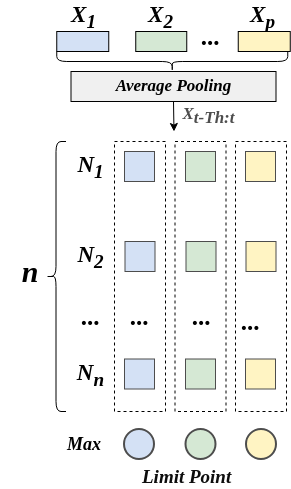} 
        \caption{}
        \label{fig4a}
    \end{subfigure}
    \begin{subfigure}[b]{0.3\textwidth}
        \centering
        \includegraphics[width=\textwidth]{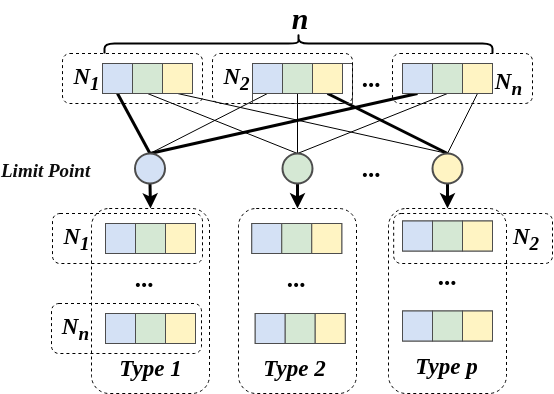} 
        \caption{}
        \label{fig4b}
    \end{subfigure}
    \caption{(a) Constructing the limit point. (b) Clustering the nodes using the Euclidean distance.}
    \label{fig4}
\end{figure}

As illustrated in Figure \ref{fig4a}, the clustering feature space is constructed based on the $P$ types of traffic patterns:
\begin{equation}
\centering
\begin{gathered}
    R_t = \frac{(X_1 W_1 || X_2 W_2 \dots || X_P W_P)}{\hat{X}_{t-T_h:t} W} \\
    R = \frac{1}{T_h} \sum_{t=1}^{T_h} R_t \\
    C_j = \text{Max}(R_j), \quad j=1,2,\dots,P \\
\end{gathered}
\end{equation}

The weight matrices \( W_i \in \mathbb{R}^{D \times 1} \) for \( i = 1, 2, \ldots, P \) transform the enriched multi-dimensional features into a single dimension. We determine the maximum value of \( R \) across the \( P \) dimensions, resulting in the limit point matrix \( C \in \mathbb{R}^P \), where \( C_p \) represents the maximum extremum state of the original traffic feature associated with feature \( P \).

\begin{equation}
\centering
    \begin{gathered}
        D_{i,j} = |R_{i,j} - C_j|, \quad j=1,2,\dots,P \\
        T_i = \text{argmin}(D_i), \quad i \in S_T \\
    \end{gathered}
\end{equation}

As shown in Figure \ref{fig4b}, the Euclidean distance matrix $D \in \mathbb{R}^{N \times P}$ is calculated between the feature tensor $R$ and the limit point matrix $C$ in the P-dimensional space. The type $T_i \in \mathbb{R}^1$ of node $i$ is the type of the limit point with the minimum Euclidean distance, placing $i$ in the node sequence pool $S_T$ for type $T_i$, which is used for subsequent node placement. The time complexity of this clustering algorithm is $O(N)$, which significantly saves computational resources compared to traditional clustering algorithms.
\subsection{DSTGG Module}
We introduce the STGG Module to model the pairwise latent spatiotemporal relationships between subsets of nodes, generating a simple subgraph containing spatiotemporal fusion relationships.
\begin{equation}
\begin{gathered}
    M_1 = \tanh(\alpha E_p^1 W_1) \\
    M_2 = \tanh(\alpha E_p^2 W_2) \\
    \hat{A}_p^s = \alpha (M_1 M_2^T - M_1^T M_2) \\
\end{gathered}
\end{equation}
The matrices $E_P^1 \in \mathbb{R}^{N_P \times D_s}$ and $E_P^2 \in \mathbb{R}^{N_P \times D_s}$ represent the node embeddings for cluster $P$ that are randomly initialized. The adjacency matrix $\hat{A}_p^s \in \mathbb{R}^{N_p \times N_p}$ containing the spatial relationships of cluster $P$ is generated.
\begin{equation}
\centering
    \begin{gathered}
        \hat{T}_p^D = (T_i^D \, ||  \, \dots \, || \, T_{i+N_p}^D), \quad \{i, \dots, i+N_p\} \subseteq V_p \\
        \hat{T}_p^W = (T_i^W \, ||  \, \dots \, || \, T_{i+N_p}^W), \quad \{i,  \dots, i+N_p\} \subseteq V_p \\
        A_p^t = \hat{T}_p^D \left(\hat{T}_p^W\right)^T \\
        \hat{A}_p^t = \beta \left( \text{ReLU} \left( \tanh \left( \frac{1}{T_h} \sum_{j=1}^{T_h} A_p^t[j,:] \right) \right) \right) \\
    \end{gathered}
\end{equation}
The matrix $A_p^t \in \mathbb{R}^{T_h \times N_p \times N_p}$ captures the periodic temporal features of daily and weekly patterns. By applying average pooling and non-linear transformations, we obtain the fused temporal information matrix $\hat{A}_p^t$ within the $T_h$ time window for cluster $P$.

\begin{equation}
    \begin{gathered}
    A_p = \text{ReLU}\left( \tanh\left( \beta \hat{A}_p^s \left(\hat{A}_p^t\right)^T \right) \right) \\
        \hat{A}_p = \begin{cases} 
      A_{i,j} & \text{if } j \in \text{argtopk}(A[i,:]) \\ 
      0 & \text{otherwise} 
   \end{cases}
    \end{gathered}
\end{equation}

 Using the sparsified spatial graph and the fused temporal information matrix, we construct a simple graph containing heterogeneous spatiotemporal information.
 \subsection{SIE Module}
This article introduces the SIE module, which is used to capture clustering features and spatiotemporal patterns in dynamic fused subgraphs, as shown in Figure \ref{fig5}.

To achieve information control and capture long-term dependencies, MHGNet introduces a gating mechanism in the graph convolution operation:
\begin{figure}[t]
    \centering
    \begin{subfigure}[b]{0.16\textwidth}
        \centering
        \includegraphics[width=\textwidth]{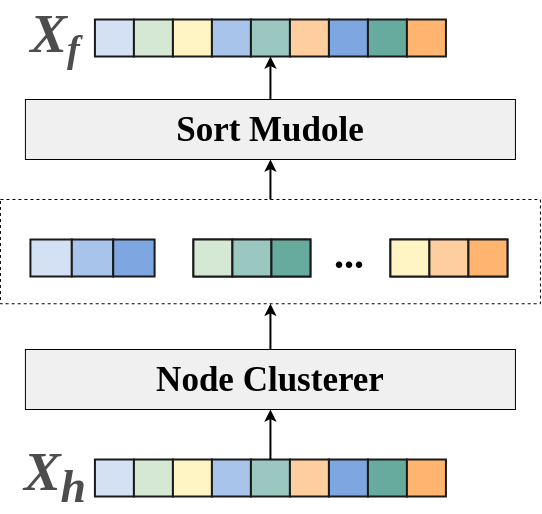} 
        \caption{}
        \label{fig5a}
    \end{subfigure}
    \begin{subfigure}[b]{0.28\textwidth}
        \centering
        \includegraphics[width=\textwidth]{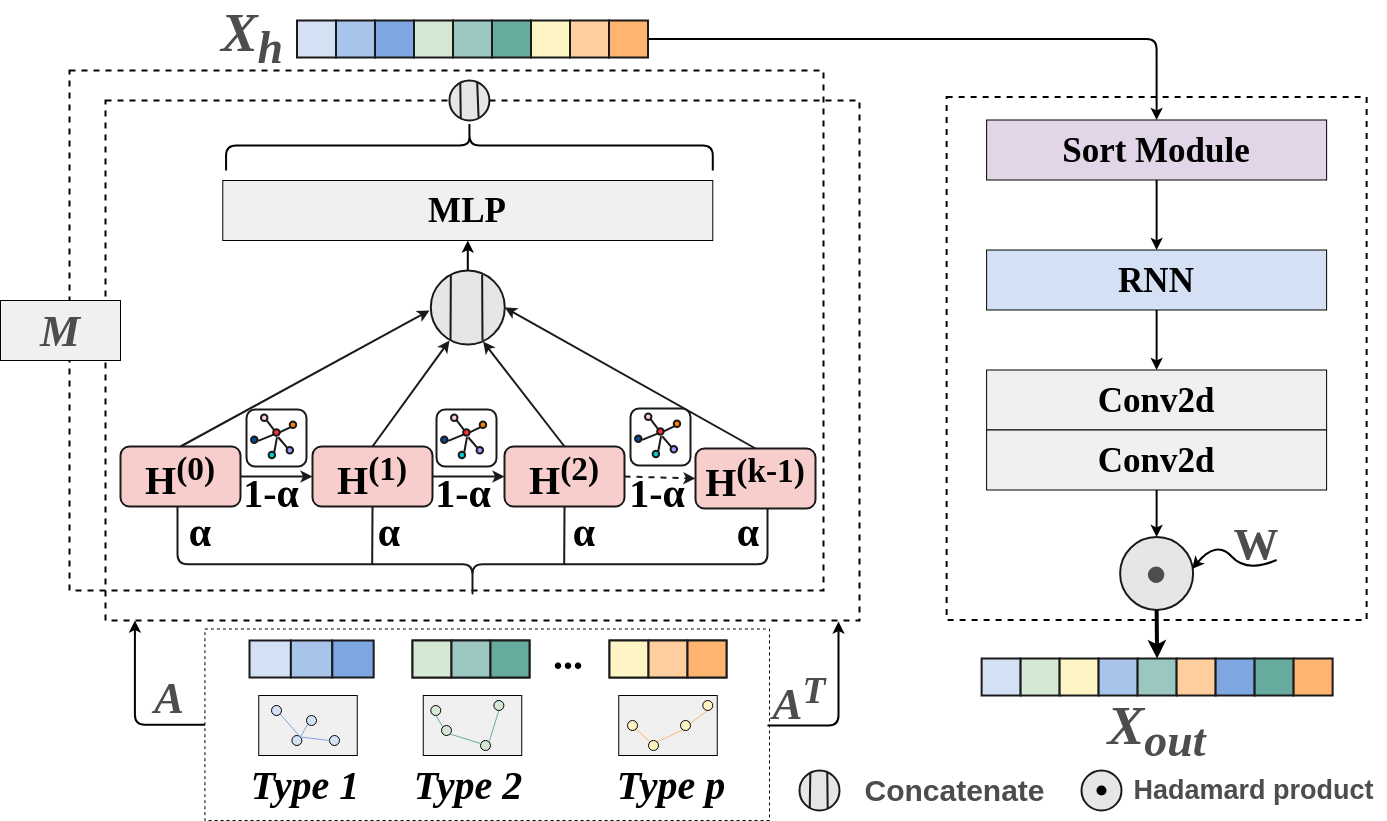} 
        \caption{}
        \label{fig5b}
    \end{subfigure}
    \caption{(a) Node clustering and node repositioning demonstration. (b) SIE Module schematic diagram. }
    \label{fig5}
\end{figure}
\begin{equation}
    \begin{gathered}
        \tilde{A}_p = \hat{A}_p + I \\
        H_p^{(k)} = \gamma H_p + (1-\gamma)(\tilde{D}^{-1} \tilde{A}) H_p^{(k-1)} \\
        \hat{H}_p = \left(H_p^{(0)} \, || \, H_p^{(1)} \, || \, \dots \, || \, H_p^{(k-1)}\right) W \\
\end{gathered}
\end{equation}
In the formulas, let $\tilde{A}_p$ denote the adjacency matrix that includes self-loops. The degree matrix $\tilde{D} \in \mathbb{R}^{N \times 1}$ is defined such that $\tilde{D}_{ii} = \sum_{j} \tilde{A}_{ij}$
for each node $i$. Additionally, the parameter $k$ represents the depth of propagation.
\begin{equation}
    \begin{gathered}
        X_h = \left(\hat{H}_1 \, || \, \hat{H}_2 \, || \, \dots \, || \, \hat{H}_p\right) \\
        \hat{X}_h[:,i,:] = X_h[:,j,:], \quad i \in S_N, \, j \in S_T \\
    \end{gathered}
\end{equation}

In these equations, $X_h$ represents the shuffled subgraph convolution results, $S_N$ denotes a continuous permutation from 0 to $N$, and $S_T$ is the node sequence pool generated by the Node Clusterer. The Sort Module regresses the node sequence, which is then used by the RNN to capture long-term temporal dependencies.

The model stacks the input over $T_c$ time steps in the GRU to form an output tensor \(H_{\text{out}} \in \mathbb{R}^{T_c \times N \times (M \times D)}\), and then performs weight redistribution on $H_{\text{out}}$:
\begin{equation}
    \begin{gathered}
         H_{\text{out}} = \text{dropout}\left([H^{(0)} \, H^{(1)} \, H^{(2)} \, \dots \, H^{(T_c-1)}]\right) \\
                 \hat{H}_{\text{out}} = \left(\text{ReLU}\left(H_{\text{out}} * W_1\right) * W_2\right) \\
        X_{\text{out}} = \hat{H}_{\text{out}} \odot W \\
    \end{gathered}
\end{equation}
\subsection{Output and prediction}
For the multi-scale spatiotemporal information captured by the SIE Module, we establish skip connections to incorporate additional feature information:
\begin{equation}
    H = \left\{X_{\text{out}} \, || \, X_{t-T_h:t} \, || \, X_1 \, || \,  \dots \, || \, X_p \, || \, T^D  \, || \, T^W\right\}
\end{equation}
The obtained tensor \(H\) is fed into the regression layer for feature integration and nonlinear transformation to obtain the prediction state \(X_{t:t+T_f}\):
\begin{equation}
    X_{t:t+T_f} = \left[ (\text{ReLU}\left( \text{ReLU}(H) W_1 \right) W_2)  * W\right]^T
\end{equation}
\section{Experiments}
\subsection{Datasets}
We used two traffic flow datasets from the California Department of Transportation (CalTrans) Performance Measurement System (PEMS): PEMS04 and PEMS08, along with two traffic speed datasets: METR-LA and PEMS-BAY. The raw data is aggregated at 5-minute time intervals and organized according to predefined spatial graphs constructed based on real road networks. For detailed information about the aforementioned datasets, see Table \ref{tab2}.
\begin{table}[htbp]
    \centering
    \caption{Dataset Statistics}
    \begin{tabular}{>{\raggedright\arraybackslash}m{1.8cm}cccc}
        \toprule[1.5pt]
        \textbf{Dataset} & \textbf{\#Nodes} & \textbf{\#Edges} & \textbf{\#Days} & \textbf{\#Data} \\
        \midrule
        PEMS04    & 307  & 340  & 59  & 16992  \\
        PEMS08    & 170  & 295  & 62  & 17856  \\
        METR-LA &207 &1722 &119 &34272 \\
        PEMS-BAY  & 325  & 2694 & 181 & 52116  \\
        \bottomrule[1.5pt]
    \end{tabular}
    \label{tab2}
\end{table}
\subsection{Baseline Methods}
\textbf{Baseline Methods}: GWNet\cite{GWNet}, DCRNN\cite{DCRNN}, STGNCDE\cite{STGNCDE}, GMAN\cite{GMAN}, AGCRN\cite{AGCRN}, STNorm\cite{STNorm}, ASTGNN\cite{ASTGNN}, STID\cite{STID}, DDGCRN\cite{DDGCRN}, PDFormer\cite{PDFormer}
\subsection{Experiment Settings}
The experiment was conducted on a system equipped with an NVIDIA GeForce RTX 4090 GPU. The METR-LA and PEMS-BAY datasets were divided into training, testing, and validation sets in a 7:2:1 ratio. For the PEMS04 and PEMS08 datasets, we used a 6:2:2 split. The model was trained using the Adam optimizer, with a weight decay (wdecay) of $1.0 \times 10^{-5}$ and an epsilon (eps) value of $1.0 \times 10^{-8}$. Implement a warm-up strategy during the first 20 training cycles, followed by curriculum learning with a length of 3. The model parameters for the four datasets are detailed in Table \ref{tab3}. The source code of MHGNet is  available: https://github.com/meiwu5/MHGNet.git

\begin{table}[h]
    \centering
    \caption{Model Parameters for Datasets}
      \resizebox{.45\textwidth}{!}{
    \begin{tabular}{cccccccc}
    \toprule[1.5pt]
    Dataset &Batch size &$D_s$ &$D_t$ &P &lr\\
    \midrule
         PEMS04&64&10 &10 &3 &0.006 \\
         PEMS08&64 &10 &10 &3 &0.006\\
         METR-LA&32 &10 &10 &2 &0.002\\
         PEMS-BAY&32 &10 &10 &2 &0.002\\
         \bottomrule[1.5pt]
    \end{tabular}
    }
    \label{tab3}
\end{table}
\subsection{Experiment Results}
\begin{table}[h]
  \centering
  \caption{The Average Metrics on the Traffic Flow Dataset.}
  \resizebox{.48\textwidth}{!}{
    \begin{tabular}{ccccccccccccc}
      \toprule[1.5pt]
      \multirow{2}{*}{Models}  & \multicolumn{3}{c}{PEMS04} & \multicolumn{3}{c}{PEMS08}\\
      \cmidrule[0.75pt]{2-7}
      & MAE & RMSE & MAPE & MAE & RMSE & MAPE\\
      \midrule[0.75pt]
      GWNet &19.36 &31.72 &13.30\%  &15.06 &24.86 &9.51\% \\
      DCRNN   &19.63 &31.26 &13.59\%   &15.22 &24.17 &10.21\% \\
      STGNCDE &19.21 &31.09 &12.76\% &15.45 &24.81 &9.92\% \\
      GMAN &19.14 &31.60 &13.19\%  &15.31 &24.92 &10.13\% \\
      AGCRN &19.38 &31.25 &13.40\% &15.32 &24.41 &10.03\% \\
      STNorm  &18.96 &30.98 &12.69\%  &15.41 &24.77 &9.76\% \\
      ASTGNN &18.60&31.03&12.63\%&14.97&24.71&9.49\% \\
      DDGCRN &18.45 &30.51 &12.19\% &14.40 &23.75 &9.40\% \\
      STID &18.38&\textbf{29.95}&12.04\%&14.21&23.28&9.27\% \\
      PDFormer &18.36 &30.03 &\textbf{12.00\%} &\textbf{13.58} &23.41 &9.05\% \\
      \midrule[0.75pt]
      \textbf{MHGNet}  &\textbf{18.32} &29.96&12.30\% 
      &13.69 &\textbf{23.22} &\textbf{9.03\%} \\
      \bottomrule[1.5pt]
    \end{tabular}
  }
  \label{tab4}
\end{table}
\begin{table}
  \centering
    \caption{Performance Metrics on the Traffic Speed Dataset}
  \resizebox{.48\textwidth}{!}{
    \begin{tabular}{c|c|ccccccccccc}
      \toprule[1.5pt]
      \multirow{2}{*}{Dataset} & \multirow{2}{*}{Models} & \multicolumn{3}{c}{Horizon 3} & \multicolumn{3}{c}{Horizon 6} & \multicolumn{3}{c}{Horizon 12} \\
      \cmidrule[0.75pt]{3-5}
      \cmidrule[0.75pt]{6-8}
      \cmidrule[0.75pt]{9-11}
       & &MAE&RMSE &MAPE &MAE &RMSE &MAPE &MAE &RMSE &MAPE \\
      \cmidrule[0.75pt]{1-11}
      \multirow{4}{*}{METR-LA}
        &AGCRN &2.87 &5.58 &7.70\% &3.23 &6.58 &9.00\% &3.62 &7.51 &10.38\% \\
      &STNorm&2.81&5.57&\textbf{7.40\%} &3.18&6.59&\textbf{8.47\%}&3.57&7.51&\textbf{10.24\%} \\
&STID&2.82&5.53&7.75\%&3.19&6.57&9.39\%&3.55&7.55&10.95\%\\
&PDFormer&2.83&\textbf{5.45}&7.77\%&3.20&\textbf{6.46}&9.19\%&3.62&\textbf{7.47}&10.91\% \\
&\textbf{MHGNet} &\textbf{2.80}&5.52&7.58\%&\textbf{3.18}&6.62&9.26\%&\textbf{3.54}&7.56&10.75\%\\
\midrule
      \multirow{4}{*}{PEMS-BAY}
      &AGCRN &1.37 &2.87 &2.94\% &1.69 &3.85 &3.87\% &1.96 &4.54 &4.64\% \\
      &STNorm &1.33 &2.82&2.76\%&1.65&3.77&3.66\%&1.92&4.45&4.46\%\\
&STID&1.31&2.79&2.78\%&1.64&3.73&3.73\%&1.91&4.42&4.55\% \\
&PDFormer & 1.32&2.83&2.78\%&1.64&3.79&3.71\%&1.91&4.43&4.51\% \\
&\textbf{MHGNet}&\textbf{1.31}&\textbf{2.74}&\textbf{2.75\%}&\textbf{1.62}&\textbf{3.65}&\textbf{3.62\%}&\textbf{1.90}&\textbf{4.36}&\textbf{4.43\%}\\
      \bottomrule[1.5pt]
    \end{tabular}
    }
    \label{tab5}
  \end{table}
As shown in Tables \ref{tab4} and \ref{tab5}, our method achieves better performance across most metrics on all four datasets. We can draw the following conclusions: Compared to GNN-based models (such as DDGCRN and AGCRN), MHGNet improves performance through subgraph-based heterogeneous graph modeling. Compared to embedding-based models like STID, it demonstrates better adaptability in handling complex traffic data. Additionally, when compared to attention-based models (such as ASTGNN and PDFormer), it exhibits good robustness and generalization performance, particularly excelling in the traffic speed dataset (such as PEMS-BAY).
\begin{table}[h]
\centering
\caption{Comparison of parameter count}
\resizebox{.48\textwidth}{!}{
\begin{tabular}{cccccc}
\toprule[1.5pt]
Dataset& DDGCRN & STID & PDFormer &$\text{MHGNet}^{\phi}$ & $\text{MHGNet}$\\
\midrule[0.75pt]
PEMS04 & 569254 & 121068 & 531165 &655268& 266470 \\
PEMS08 & 311759 & 116684 & 531165 &303139& 166332 \\
\bottomrule[1.5pt]
\end{tabular}%
}
\label{tab6}
\end{table}
\begin{itemize}
    \item $\textbf{MHGNet}^{\phi}$: The variant with two layers of whole-graph convolution after removing the node classifier.
\end{itemize}
\subsection{Ablation Experiments}
\begin{table}[h]
  \centering
  \caption{Node Clusterer Ablation}
  \resizebox{.48\textwidth}{!}{
    \begin{tabular}{ccccccccccc}
      \toprule[1.5pt]
      \multirow{2}{*}{Dataset} & \multirow{2}{*}{Variants} & \multicolumn{3}{c}{Horizon 3} &\multicolumn{3}{c}{Horizon 6} &\multicolumn{3}{c}{Horizon 12}  \\
      \cmidrule{3-11}
      & & MAE & RMSE & MAPE & MAE & RMSE & MAPE & MAE & RMSE & MAPE \\
      \cmidrule[0.75pt]{1-11}
      \multirow{3}{*}{PEMS04} 
      & w/o Node Clusterer & 17.81 &28.89 &11.86\% & 18.70 &30.36 &12.37\% & 20.30 &32.61 &13.42\% \\
      & $\text{MHGNet}_{p=2}$  &17.57 &28.76 &\textbf{11.69\%} &
      18.43 &30.17 &12.15\% &19.78 &32.18 &13.18\% \\
      & $\text{MHGNet}_{p=3}$ &\textbf{17.48} &\textbf{28.63} &11.73\% & \textbf{18.30} &\textbf{30.03} &\textbf{12.06\%} & \textbf{19.69} &\textbf{32.07} &\textbf{12.96\%} \\
      \cmidrule[0.75pt]{1-11}
      \multirow{3}{*}{PEMS08} 
      & w/o Node Clusterer &13.81 &22.08 &9.00\% &14.61 &23.82 &9.53\% &16.09 &26.38 &10.57\% \\
      & $\text{MHGNet}_{p=2}$ &12.84 &\textbf{21.54} &\textbf{8.41\%} &13.69 &\textbf{23.35} &\textbf{8.99\%} &15.76 &26.06 &10.29\% \\
      & $\text{MHGNet}_{p=3}$ &\textbf{12.77} &\textbf{21.54} &8.47\% & \textbf{13.61} &23.36 &9.03\% & \textbf{15.58} &\textbf{25.93} &\textbf{10.20\%} \\
            \cmidrule[0.75pt]{1-11}
      \multirow{3}{*}{METR-LA} 
      & w/o Node Clusterer &2.83 &5.53 &7.67\% &3.22 &6.61 &9.46\% &3.76 &7.77 &11.67\% \\
      & $\text{MHGNet}_{p=2}$ &\textbf{2.80}&\textbf{5.52}&\textbf{7.58\%}&3.18&6.62&\textbf{9.26\%}&\textbf{3.54}&\textbf{7.56}&\textbf{10.75\%} \\
      & $\text{MHGNet}_{p=3}$ &\textbf{2.80} &5.56 &7.59\% &\textbf{3.17} &\textbf{6.58} &9.29\% &3.56 &7.58 &10.98\% \\
                  \cmidrule[0.75pt]{1-11}
      \multirow{3}{*}{PEMS-BAY} 
      & w/o Node Clusterer &1.34 &2.82 &2.87\% &1.66 &3.77 &3.74\% &1.97 &4.43 &4.65\% \\
      & $\text{MHGNet}_{p=2}$ &\textbf{1.31} &\textbf{2.74} &\textbf{2.75\%} &\textbf{1.62} &\textbf{3.65} &\textbf{3.62\%} &\textbf{1.90} &\textbf{4.36} &\textbf{4.43\%} \\
      & $\text{MHGNet}_{p=3}$ &1.32 &2.79 &2.81\% &1.63 &3.72 &3.70\% &1.92 &4.42 &4.53\% \\
      \bottomrule[1.5pt]
    \end{tabular}
    \label{tab7}
   }
\end{table}
\begin{itemize}
    \item \textbf{w/o Node Clusterer}: No node clustering, perform a single graph convolution
    \item \textbf{$\text{MHGNet}_{p=2}$}: Node clustering into 2 patterns followed by subgraph convolution
    \item \textbf{$\text{MHGNet}_{p=3}$}: Node clustering into 3 patterns followed by subgraph convolution
\end{itemize}
\begin{table}[h]
  \centering
  \caption{Spatiotemporal Graph Ablation.}
  \resizebox{.48\textwidth}{!}{
    \begin{tabular}{ccccccccccccc}
      \toprule[1.5pt]
      \multirow{2}{*}{Models}  & \multicolumn{3}{c}{PEMS04} & \multicolumn{3}{c}{PEMS08}\\
      \cmidrule[0.75pt]{2-7}
      & MAE & RMSE & MAPE & MAE & RMSE & MAPE\\
      \midrule[0.75pt]
      w/o sg &18.73 &30.31 &12.67\% &14.07 &23.43 &9.19\% \\
      w/o tg &18.56 &30.09 &12.97\% &14.08 &23.50 &9.29\% \\
      $\text{MHGNet}_{p=2}$ &18.41 &30.08 &12.20\% &13.81 &23.31 &9.08\% \\
      \bottomrule[1.5pt]
    \end{tabular}
  }
  \label{tab8}
\end{table}
\begin{itemize}
    \item \textbf{w/o sg}: Removing $\hat{A}_p^s$ in the DSTGNN Module.
    \item \textbf{w/o tg}: Removing $\hat{A}_p^t$in the DSTGNN Module.
    \item \textbf{$\text{MHGNet}_{p=2}$}: Using the spatiotemporal fused graph $\hat{A}_p$.
\end{itemize}

\section{Conclusion}
In this paper, we propose MHGNet, a new framework for modeling spatiotemporal heterogeneous graphs. The model utilizes a timestamp embedding matrix and a node embedding matrix to decouple one-dimensional features into a multi-dimensional feature space, enabling simple clustering. During the clustering process, a dynamic graph containing spatiotemporal information is generated for subgraph convolution. Moreover, experiments are conducted on multiple datasets.


\begin{thebibliography}{00}
\bibitem{conditions} D. A. Tedjopurnomo, Z. Bao, B. Zheng, F. M. Choudhury, and A. K. Qin, ``A Survey on Modern Deep Neural Network for Traffic Prediction: Trends, Methods and Challenges,'' \textit{IEEE Transactions on Knowledge and Data Engineering}, vol. 34, no. 4, pp. 1544--1561, 2022. [DOI: 10.1109/TKDE.2020.3001195].

\bibitem{conditions2} W. Shao \textit{et al.}, ``Long-term Spatio-Temporal Forecasting via Dynamic Multiple-Graph Attention,'' in \textit{Proceedings of the Thirty-First International Joint Conference on Artificial Intelligence}, pp. 2225--2232, 2022. [DOI: 10.24963/ijcai.2022/309].

\bibitem{conditions3} Y. Lv, Y. Duan, W. Kang, Z. Li, and F.-Y. Wang, ``Traffic Flow Prediction With Big Data: A Deep Learning Approach,'' \textit{IEEE Transactions on Intelligent Transportation Systems}, vol. 16, no. 2, pp. 865--873, 2015. [DOI: 10.1109/TITS.2014.2345663].

\bibitem{research1} J. S. Gracias, G. S. Parnell, E. Specking, E. A. Pohl, and R. Buchanan, ``Smart Cities—A Structured Literature Review,'' \textit{Smart Cities}, vol. 6, no. 4, pp. 1719--1743, 2023. [DOI: 10.3390/smartcities6040080].

\bibitem{research2} Z. Liu, Z. Li, K. Wu, and M. Li, ``Urban Traffic Prediction from Mobility Data Using Deep Learning,'' \textit{IEEE Network}, vol. 32, no. 4, pp. 40--46, 2018. [DOI: 10.1109/MNET.2018.1700411].

\bibitem{research3} Z. Pan \textit{et al.}, ``Urban Traffic Prediction from Spatio-Temporal Data Using Deep Meta Learning,'' in \textit{Proceedings of the 25th ACM SIGKDD International Conference on Knowledge Discovery \& Data Mining}, pp. 1720--1730, 2019. [DOI: 10.1145/3292500.3330884].

\bibitem{prediction} A. B. P and R. Sumathi, ``Data Sources for Urban Traffic Prediction: A Review on Classification, Comparison and Technologies,'' in \textit{2020 3rd International Conference on Intelligent Sustainable Systems (ICISS)}, pp. 628--635, 2020. [DOI: 10.1109/ICISS49785.2020.9316096].

\bibitem{prediction2} Z. Pan \textit{et al.}, ``Spatio-Temporal Meta Learning for Urban Traffic Prediction,'' \textit{IEEE Transactions on Knowledge and Data Engineering}, vol. 34, no. 3, pp. 1462--1476, 2022. [DOI: 10.1109/TKDE.2020.2995855].

\bibitem{GCN} F. Scarselli \textit{et al.}, ``The Graph Neural Network Model,'' \textit{IEEE Transactions on Neural Networks}, vol. 20, no. 1, pp. 61--80, 2009. [DOI: 10.1109/TNN.2008.2005605].

\bibitem{GCN2} X. Wang \textit{et al.}, ``Traffic Flow Prediction via Spatial Temporal Graph Neural Network,'' in \textit{Proceedings of The Web Conference 2020}, pp. 1082--1092, 2020. [DOI: 10.1145/3366423.3380186].

\bibitem{tasks} H. Yao \textit{et al.}, ``Revisiting Spatial-Temporal Similarity: A Deep Learning Framework for Traffic Prediction,'' \textit{Proceedings of the AAAI Conference on Artificial Intelligence}, vol. 33, no. 1, pp. 5668--5675, 2019. [DOI: 10.1609/aaai.v33i01.33015668].

\bibitem{tasks2} T. Liu \textit{et al.}, ``Graph-Based Dynamic Modeling and Traffic Prediction of Urban Road Network,'' \textit{IEEE Sensors Journal}, vol. 21, no. 24, pp. 28118--28130, 2021. [DOI: 10.1109/JSEN.2021.3124818].

\bibitem{AspEm} Y. Shi \textit{et al.}, ``AspEm: Embedding Learning by Aspects in Heterogeneous Information Networks,'' 2018. [Online]. Available: https://arxiv.org/abs/1803.01848.

\bibitem{HEER} Y. Shi \textit{et al.}, ``Easing Embedding Learning by Comprehensive Transcription of Heterogeneous Information Networks,'' in \textit{Proceedings of the 24th ACM SIGKDD International Conference on Knowledge Discovery \& Data Mining}, pp. 2190--2199, 2018. [DOI: 10.1145/3219819.3220006].

\bibitem{COMPGCN} S. Vashishth \textit{et al.}, ``Composition-based Multi-Relational Graph Convolutional Networks,'' 2020. [Online]. Available: https://arxiv.org/abs/1911.03082.

\bibitem{networks} X. Yin \textit{et al.}, ``Deep Learning on Traffic Prediction: Methods, Analysis, and Future Directions,'' \textit{IEEE Transactions on Intelligent Transportation Systems}, vol. 23, no. 6, pp. 4927--4943, 2022. [DOI: 10.1109/TITS.2021.3054840].

\bibitem{networks2} X. Shi \textit{et al.}, ``A Spatial–Temporal Attention Approach for Traffic Prediction,'' \textit{IEEE Transactions on Intelligent Transportation Systems}, vol. 22, no. 8, pp. 4909--4918, 2021. [DOI: 10.1109/TITS.2020.2983651].

\bibitem{networks3} M. Shaygan \textit{et al.}, ``Traffic prediction using artificial intelligence: Review of recent advances and emerging opportunities,'' \textit{Transportation Research Part C: Emerging Technologies}, vol. 145, p. 103921, 2022. [DOI: 10.1016/j.trc.2022.103921].

\bibitem{GWNet} Z. Wu \textit{et al.}, ``Graph WaveNet for Deep Spatial-Temporal Graph Modeling,'' 2019. [Online]. Available: https://arxiv.org/abs/1906.00121.

\bibitem{DCRNN} Y. Li, R. Yu, C. Shahabi, and Y. Liu, ``Diffusion Convolutional Recurrent Neural Network: Data-Driven Traffic Forecasting,'' in \textit{International Conference on Learning Representations}, 2018. [Online]. Available: https://openreview.net/forum?id=SJiHXGWAZ.

\bibitem{STGNCDE} J. Choi and N. Park, ``Graph Neural Rough Differential Equations for Traffic Forecasting,'' \textit{ACM Trans. Intell. Syst. Technol.}, vol. 14, no. 4, art. 74, 2023. [DOI: 10.1145/3604808].

\bibitem{GMAN} C. Zheng \textit{et al.}, ``GMAN: A Graph Multi-Attention Network for Traffic Prediction,'' 2019. [Online]. Available: https://arxiv.org/abs/1911.08415.


\bibitem{AGCRN} L. Bai \textit{et al.}, ``Adaptive Graph Convolutional Recurrent Network for Traffic Forecasting,'' 2020. [Online]. Available: https://arxiv.org/abs/2007.02842.

\bibitem{STNorm} J. Deng \textit{et al.}, ``ST-Norm: Spatial and Temporal Normalization for Multi-variate Time Series Forecasting,'' in \textit{Proceedings of the 27th ACM SIGKDD Conference on Knowledge Discovery \& Data Mining}, pp. 269–278, 2021. [DOI: 10.1145/3447548.3467330].

\bibitem{ASTGNN} S. Guo \textit{et al.}, ``Learning Dynamics and Heterogeneity of Spatial-Temporal Graph Data for Traffic Forecasting,'' \textit{IEEE Transactions on Knowledge and Data Engineering}, vol. 34, no. 11, pp. 5415--5428, 2022. [DOI: 10.1109/TKDE.2021.3056502].

\bibitem{STID} Z. Shao \textit{et al.}, ``Spatial-Temporal Identity: A Simple yet Effective Baseline for Multivariate Time Series Forecasting,'' in \textit{Proceedings of the 31st ACM International Conference on Information \& Knowledge Management}, pp. 4454--4458, 2022. [DOI: 10.1145/3511808.3557702].

\bibitem{DDGCRN} W. Weng \textit{et al.}, ``A Decomposition Dynamic Graph Convolutional Recurrent Network for Traffic Forecasting,'' \textit{Pattern Recognition}, vol. 142, p. 109670, 2023. [DOI: 10.1016/j.patcog.2023.109670].


\bibitem{PDFormer} J. Jiang \textit{et al.}, ``PDFormer: Propagation Delay-Aware Dynamic Long-Range Transformer for Traffic Flow Prediction,'' 2024. [Online]. Available: https://arxiv.org/abs/2301.07945.

\end{thebibliography}
\end{document}